\documentclass[10pt,twocolumn,letterpaper]{article}

\usepackage{cvpr}      
\definecolor{cvprblue}{rgb}{0.21,0.49,0.74}
\usepackage[pagebackref,breaklinks,colorlinks,allcolors=cvprblue]{hyperref}

\def\paperID{46} 
\def\confName{CVPR}
\def\confYear{2026}

\title{Few-Shot Synthetic Data Generation with Diffusion Models for Downstream Vision Tasks}

\author{
Daniil Dushenev$^{1}$ \and
Nazariy Karpov$^{1}$ \and
Daniil Zinovjev$^{1}$ \and
Alexander Gorin$^{1}$ \and
Konstantin Kulikov$^{1}$\\[0.5em]
$^{1}$National University of Science and Technology MISIS, Moscow, Russia
}

\begin{document}
\maketitle
\begin{abstract}

Class imbalance is a persistent challenge in visual recognition, particularly in safety-critical domains where collecting positive examples is expensive and \textbf{rare events} are inherently underrepresented. 
We propose a lightweight synthetic data augmentation pipeline that fine-tunes a LoRA adapter on as few as \textbf{20--50} real images of a rare class and uses a pretrained diffusion model to generate synthetic samples for training.

We systematically vary the synthetic-to-real ratio and evaluate the approach across two structurally different domains: chest X-ray pathology classification (NIH ChestX-ray14) and industrial surface crack detection (Magnetic Tile Defect dataset). 
All evaluations are performed on held-out sets of real images only.

Across both domains, synthetic augmentation consistently improves rare-class recall and F1 compared to training with real data alone. 
Performance improves with moderate synthetic augmentation and shows diminishing returns as the \textbf{synthetic ratio} increases.

These results suggest that LoRA-adapted diffusion models provide a simple and \textbf{scalable mechanism} for augmenting rare classes, enabling effective learning in data-scarce scenarios \textbf{across heterogeneous visual domains.}

\end{abstract}

\begin{figure}[t]
\centering
\includegraphics[width=\linewidth]{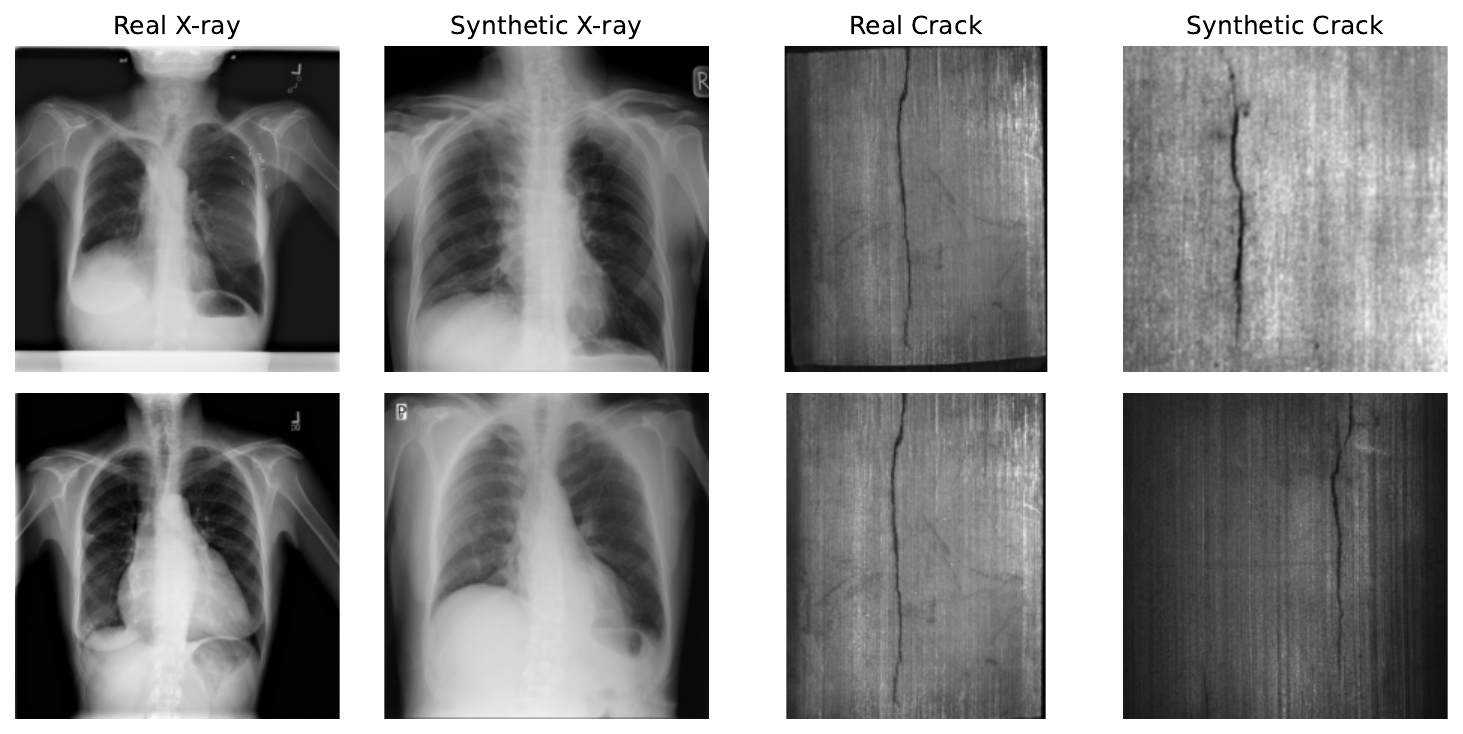}
\caption{
Examples of real and synthetic images for both datasets.
Left: chest X-ray pathology samples from the NIH ChestX-ray dataset.
Right: crack defect samples from the Magnetic Tile Defect dataset.
Synthetic images are generated using the proposed LoRA-adapted diffusion model.
}
\label{fig:qualitative}
\end{figure} 
\section{Introduction}

Class imbalance remains a major obstacle in many real-world computer vision systems. 
In safety-critical and industrial domains, collecting positive examples of rare events is often difficult or prohibitively expensive. 
Examples include medical pathologies that appear infrequently in clinical data and manufacturing defects that occur sporadically during production. 
As a result, models trained on such datasets often suffer from poor recall for the rare class.

Data augmentation is a common strategy to mitigate this problem, but traditional augmentation methods are limited to simple geometric or photometric transformations and cannot introduce fundamentally new visual variations. 
Recent advances in generative models, particularly diffusion models, offer the possibility of synthesizing realistic images that expand the diversity of training data. 
However, training generative models typically requires large datasets and substantial computational resources, making them difficult to apply in few-shot scenarios.

In this work, we investigate whether diffusion models adapted with lightweight LoRA fine-tuning can generate useful synthetic data from extremely small datasets. 
Starting from only 20--50 real images of a rare class, we train a LoRA adapter for a pretrained diffusion model and generate additional synthetic samples that are used to augment the training set for a downstream classifier.

To evaluate the generality of this approach, we study two structurally different domains: medical chest X-ray pathology classification and industrial surface defect detection. 
Despite substantial differences in modality, visual structure, and data distribution, both tasks share a common challenge: severe class imbalance and limited availability of positive examples.

Our experiments show that synthetic augmentation consistently improves rare-class detection performance in both domains. 
Furthermore, by varying the amount of generated data, we observe a predictable relationship between the synthetic-to-real ratio and classification performance, suggesting that the proposed approach provides a scalable solution for rare-class learning in data-scarce visual domains.

\textbf{Our contributions are summarized as follows:}

\begin{itemize}

\item We propose a lightweight synthetic data augmentation pipeline based on LoRA-adapted diffusion models that operates with as few as 20--50 real training images.

\item We systematically analyze the effect of synthetic dataset size by varying the synthetic-to-real ratio from $0.5\times$ to $20\times$.

\item We demonstrate that the proposed approach consistently improves rare-class detection across two structurally different domains: medical imaging and industrial defect inspection.

\item Our results suggest that diffusion-based synthetic augmentation provides a scalable strategy for learning from rare events in data-scarce visual domains.

\end{itemize}
\section{Related Work}

\paragraph{Diffusion-based data augmentation.}
Diffusion-based generative models have recently been explored not only for image synthesis but also for semantic data augmentation. 
Unlike traditional augmentation methods that rely on geometric or photometric transformations, diffusion models enable the generation of semantically meaningful variations that can improve downstream recognition performance. 
Recent work has shown that pretrained text-to-image diffusion models can improve classification by generating synthetic training samples or performing semantic edits of existing images \cite{trabucco2023diffusion, azizi2023synthetic}. 
These results suggest that diffusion models can act as powerful data augmentation mechanisms, particularly when labeled data is scarce.

\paragraph{Few-shot diffusion adaptation.}
Adapting large generative models to new concepts from only a few examples has become feasible through parameter-efficient tuning techniques. 
LoRA introduces low-rank parameter updates that enable efficient adaptation of large pretrained models while updating only a small subset of parameters \cite{hu2021lora}. 
Building on this idea, DreamBooth and Custom Diffusion demonstrate that text-to-image diffusion models can be specialized to new visual concepts using only a handful of reference images \cite{ruiz2022dreambooth, kumari2023customdiffusion}. 
These methods make diffusion models particularly attractive for few-shot synthetic data generation.

\paragraph{Synthetic data in domain-specific applications.}
Synthetic data generation has also been explored in domains closely related to ours. 
In medical imaging, diffusion models have been used to generate chest X-rays and dermatology images to improve downstream training in data-limited settings \cite{hamamci2023roentgen, sagers2023medical}. 
Similarly, in industrial inspection, recent work has proposed diffusion-based pipelines to synthesize defect images and associated labels for training detection or segmentation models \cite{valvano2024industrial, simoni2025defect}. 
These studies highlight the potential of generative models for addressing data scarcity in specialized visual domains.

\paragraph{Evaluating synthetic data utility.}
Beyond visual realism, recent work emphasizes evaluating synthetic data by its utility for downstream tasks rather than solely by perceptual quality. 
Several studies analyze the relationship between synthetic and real data in terms of utility, fidelity, and robustness, demonstrating that synthetic data can significantly improve performance when used as training augmentation \cite{xing2023utility, agnihotri2025synthetic}. 
Our work follows this perspective and evaluates synthetic samples by their impact on downstream classification performance in real-world tasks.

\paragraph{Positioning of this work.}
While prior research has explored diffusion models for data augmentation and domain-specific synthetic data generation, most studies either focus on large-scale datasets or specialized application domains. 
In contrast, our work investigates whether \emph{few-shot diffusion adaptation} can serve as a practical augmentation strategy when only a handful of positive examples are available. 
Specifically, we study the utility of LoRA-adapted diffusion models for rare-class augmentation and analyze how performance scales with increasing amounts of synthetic data.

Furthermore, we evaluate this approach across two structurally different visual domains—medical imaging and industrial defect inspection—to assess its generality. 
This cross-domain analysis highlights the potential of diffusion-based synthetic augmentation as a scalable solution for learning from rare events in data-scarce visual recognition tasks.
\section{Method}

We study whether LoRA-adapted text-to-image diffusion models can serve as a 
practical rare-class augmentation strategy when only a handful of positive 
examples are available. Our pipeline consists of three stages: adapter fine-tuning, 
synthetic generation, and downstream classifier evaluation. We validate the 
approach on two domains---industrial surface inspection and medical imaging---using 
approximately 50 real positive-class images in each case.

\subsection{LoRA Fine-Tuning}

For each rare target class we fine-tune a pretrained text-to-image diffusion model
(FLUX.2-dev~\cite{flux2dev}) using DreamBooth-style LoRA adaptation 
\cite{ruiz2022dreambooth, hu2021lora}. Adaptation is applied to the DiT transformer 
backbone; both text encoders are kept frozen. We use rank $r{=}64$, $\alpha{=}8$, 
dropout $0.08$, and train for 200 steps with 8-bit AdamW at a learning rate of 
$5 \times 10^{-3}$ in bf16 mixed precision. To fit within a single A100 80\,GB GPU 
the transformer is NF4-quantized during training. Every training image shares a 
fixed caption of the form \texttt{"a photo of <class name>"}. No post-hoc filtering 
or manual curation is applied to the generated images.

\subsection{Synthetic Data Generation}

From each fine-tuned adapter we sample ${\sim}1{,}000$ synthetic images of the 
target rare class at $512 \times 512$ resolution using 20--24 denoising steps 
with a classifier-free guidance scale of 1.5--2.5. 
Diversity arises solely from stochastic sampling with varying random seeds; 
we do not employ prompt augmentation or multiple caption templates.

Since the LoRA adapter is trained exclusively on images of the rare class, the 
diffusion model generates samples corresponding only to that class. These 
synthetic images are therefore used to augment the positive class of the 
downstream training dataset.

\subsection{Downstream Evaluation}

We train a ResNet-18~\cite{he2016deep} (ImageNet-pretrained) with a single-output 
binary head optimised with BCEWithLogitsLoss. The positive-class weight is set 
automatically as the inverse class ratio to account for imbalance. From the pool 
of ${\sim}1{,}000$ synthetic images we construct training sets at six 
synthetic-to-real ratios --- $0.5\times$, $1\times$, $2\times$, $4\times$, $10\times$, 
and $20\times$ relative to the number of real positives---alongside a 
\emph{no-synth} baseline that uses only the ${\sim}50$ real positives. 
All conditions share an identical held-out test set. We report F1 (positive class), 
PR-AUC, and Recall averaged over 5-fold cross-validation.
\section{Experiments}

\begin{table}[t]
\centering
\caption{Effect of synthetic augmentation ratio on classification performance. 
Results are reported on test sets consisting only of real images.}
\label{tab:main_results}
\begin{tabular}{lcccc}
\toprule
Dataset & Synth Ratio & F1 & PR-AUC & Recall \\
\midrule

Magnetic Tiles & $0\times$ & 0.051 & 0.141 & 0.063 \\
               & $0.5\times$ & 0.229 & 0.236 & 0.234 \\
               & $1\times$ & 0.242 & 0.269 & 0.450 \\
               & $2\times$ & 0.277 & 0.259 & 0.405 \\
               & $4\times$ & 0.296 & 0.313 & 0.423 \\
               & $10\times$ & 0.251 & 0.304 & 0.495 \\
               & $20\times$ & 0.235 & 0.311 & \textbf{0.658} \\
               & \textbf{\textit{synth-only (20$\times$)}} & \textbf{0.310} & \textbf{0.375} & 0.614 \\

\midrule

Chest X-ray & $0\times$ & 0.193 & 0.846 & 0.130 \\
            & $0.5\times$ & 0.459 & 0.767 & 0.412 \\
            & $1\times$ & 0.407 & \textbf{0.848} & 0.318 \\
            & $2\times$ & 0.430 & 0.796 & 0.370 \\
            & $\mathbf{4\times}$ & \textbf{0.686} & 0.677 & \textbf{0.744} \\
            & $10\times$ & 0.615 & 0.738 & 0.570 \\
            & $20\times$ & 0.537 & 0.741 & 0.540 \\
            & \textit{synth-only (20$\times$)} & 0.439 & 0.738 & 0.465 \\

\bottomrule
\end{tabular}
\end{table}

\begin{figure}[t]
\centering
\includegraphics[width=0.9\linewidth]{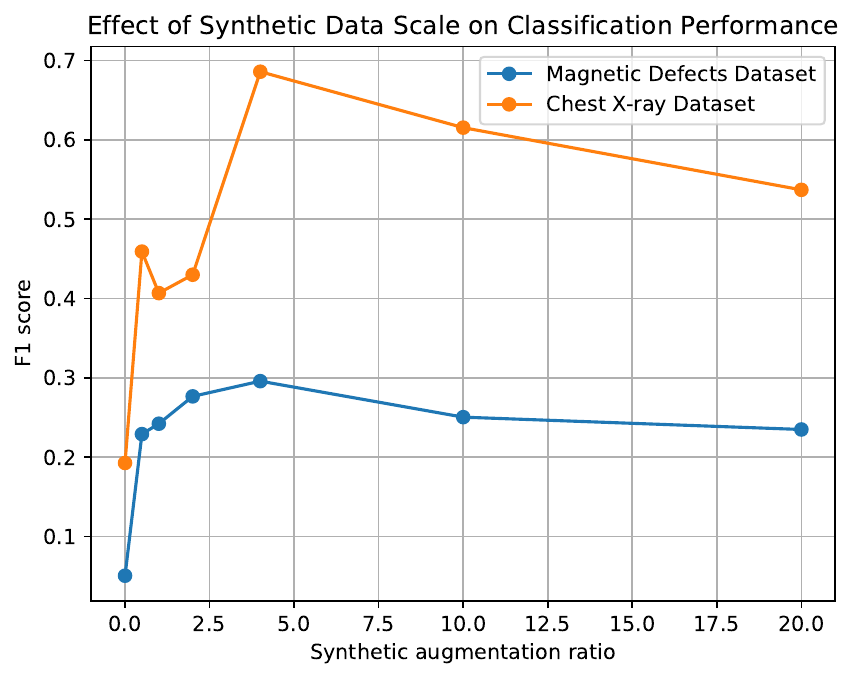}
\caption{Effect of synthetic dataset size on classification performance.
Moderate synthetic augmentation improves F1-score on both datasets,
while excessive augmentation leads to diminishing returns.}
\label{fig:f1_scaling}
\end{figure}

\subsection{Datasets}

We evaluate the proposed synthetic data augmentation approach on two datasets with severe class imbalance.

\textbf{NIH Chest X-ray Dataset.}
We use the publicly available NIH ChestX-ray14 dataset~\cite{wang2017chestxray}, which contains chest radiographs annotated with multiple pathologies.
In our experiments we focus on a binary classification task for a rare pathology.
The dataset contains a large number of negative samples and a small number of positive samples.

For diffusion adaptation, we use only a small subset of approximately
50 real images of the target pathology to train the LoRA adapter.
These images represent the rare class used for few-shot synthetic
data generation.

\textbf{Magnetic Tile Surface Defect Dataset.}
We also evaluate our method on the Magnetic Tile Surface Defect dataset, which contains images of industrial magnetic tiles with several defect categories.
We focus on the \textit{crack} defect class and formulate a binary classification task (crack vs. non-crack).
Similar to the medical dataset, the crack class is relatively rare.

For both datasets, we use a small set of real positive samples (20--50 images) to train a LoRA adapter for a diffusion model and generate synthetic images for the rare class.

\subsection{Experimental Setup}

To generate synthetic images, we fine-tune a diffusion model using a LoRA adapter trained on a small set of real images of the rare class. 
The adapted model is then used to generate additional synthetic samples that augment the training dataset.

For the downstream task, we train a ResNet18 classifier on datasets augmented with varying amounts of synthetic data. 
We evaluate synthetic-to-real ratios of $0\times$, $0.5\times$, $1\times$, $2\times$, $4\times$, $10\times$, and $20\times$, 
where the ratio indicates the number of generated samples relative to the number of real positive images.

\subsection{Metrics and Evaluation Protocol}

All evaluations are performed on test sets consisting only of real images.
This ensures that synthetic samples are used only for training and do not influence evaluation.

Due to the strong class imbalance, we report the following metrics:

\begin{itemize}
\item F1-score for the positive class
\item Precision-Recall AUC (PR-AUC)
\item Recall
\end{itemize}

F1-score is used as the primary metric because it balances precision and recall in the presence of severe class imbalance.

\subsection{Diversity analysis}
We analyze the diversity of generated samples using LPIPS and PSNR
distributions computed between pairs of images. PSNR, a pixel-level
similarity metric, shows a distribution close to that of real images,
suggesting that the generated samples preserve the overall low-level
structure of the data.

In contrast, LPIPS distributions exhibit a slight shift relative to
real images while maintaining a similar overall shape. Since LPIPS
captures perceptual similarity in a deep feature space, this shift
indicates that synthetic samples introduce semantic variation while
remaining consistent with the visual characteristics of the rare
class.

\begin{figure}[t]
\centering
\includegraphics[width=\linewidth]{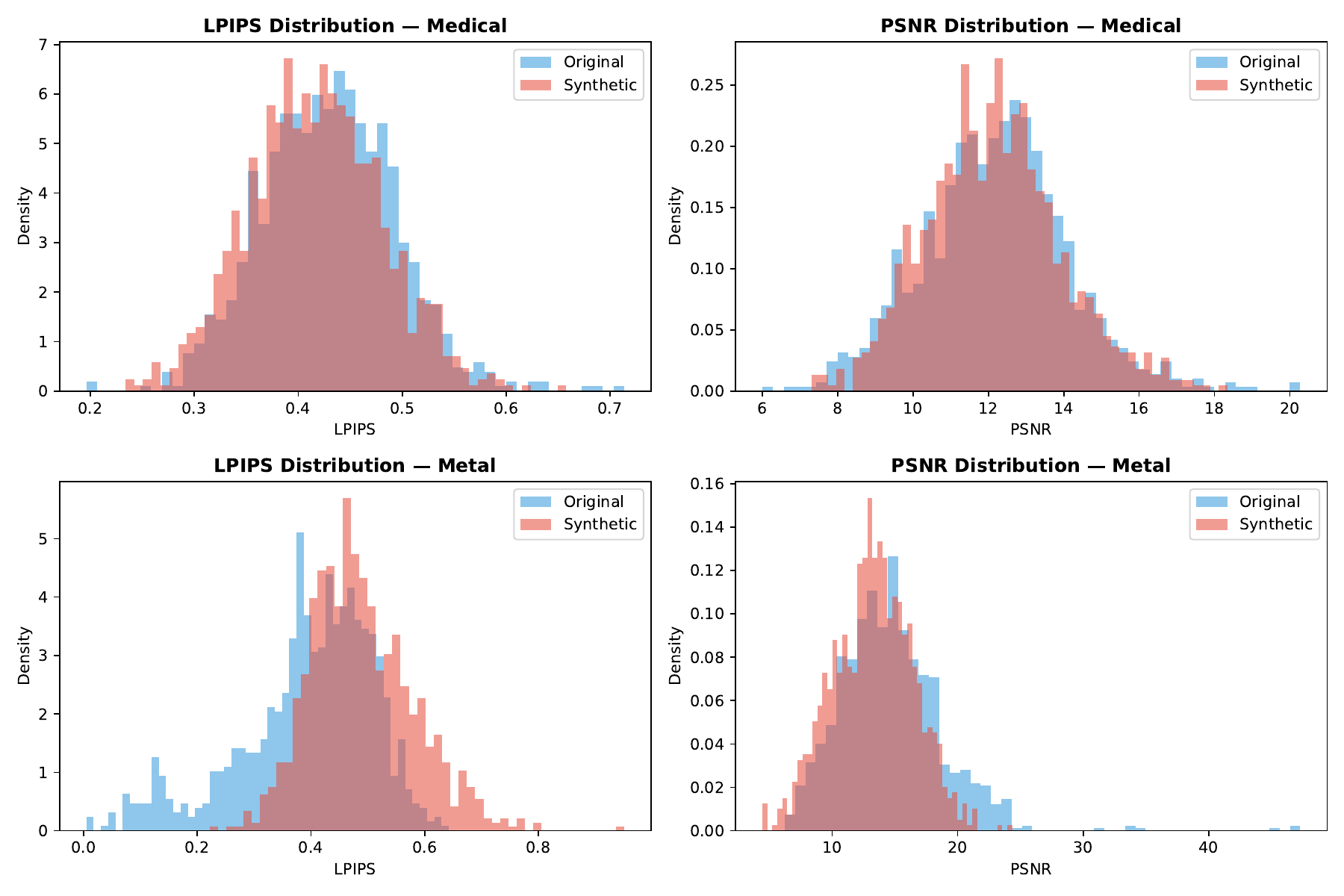}
\caption{
LPIPS and PSNR distributions for real and synthetic rare-class samples.
PSNR remains close to the real-data distribution, while LPIPS shows a
small perceptual shift with similar overall variation, indicating
diverse synthetic samples without mode collapse.
}
\label{fig:diversity_metrics}
\end{figure}

Figure~\ref{fig:diversity_metrics} compares the LPIPS and PSNR
distributions of real and synthetic rare-class samples for both
datasets.

Together, these observations suggest that the diffusion model produces
diverse samples without exhibiting mode collapse, while preserving
the structural properties of the original dataset.

\section{Results}

Table~\ref{tab:main_results} summarizes the classification performance for different synthetic augmentation ratios on both datasets. 
Training with synthetic samples consistently improves rare-class detection compared to the real-data-only baseline.

On the Magnetic Tile defect dataset, the baseline model trained without synthetic data achieves an F1-score of 0.051. 
Introducing synthetic samples significantly improves performance, with the best result of 0.296 obtained at a $4\times$ synthetic augmentation ratio. 
This corresponds to nearly a six-fold improvement over the baseline. 
Recall increases substantially as the amount of synthetic data grows, reaching 0.658 at $20\times$, indicating improved detection of the rare defect class.

A similar trend is observed on the NIH Chest X-ray dataset. 
The real-only baseline achieves an F1-score of 0.193, while synthetic augmentation increases performance up to 0.686 at a $4\times$ ratio. 
This demonstrates that synthetic samples generated from only a small number of real images can significantly improve downstream classification performance even in highly imbalanced medical datasets.

We additionally evaluate a \textbf{synthetic-only} setting where the classifier
is trained using generated rare-class samples without any real
positives. While this setup achieves reasonable recall, overall
performance remains lower than mixed real+synthetic training,
confirming that synthetic data is most effective as augmentation
rather than a replacement for real data.

Figure~\ref{fig:f1_scaling} further illustrates the relationship between the synthetic augmentation ratio and model performance. 
Moderate amounts of synthetic data lead to the largest improvements, while excessive augmentation results in diminishing returns or slight performance degradation. 
This suggests that while synthetic data improves rare-class representation, too many generated samples may distort the training distribution.

Figure~\ref{fig:diversity_metrics} shows LPIPS and PSNR distributions
for real and synthetic samples. Similar PSNR distributions indicate
preserved pixel-level structure, while LPIPS exhibits a small
perceptual shift with comparable variation. This suggests that the
generated samples remain diverse and do not exhibit mode collapse.

Overall, the results demonstrate that LoRA-adapted diffusion models provide an effective strategy for augmenting rare classes in data-scarce scenarios across structurally different visual domains.

{
    \small
    \bibliographystyle{ieeenat_fullname}
    \bibliography{main}
}


\end{document}